\documentclass[a4paper, 10 pt, conference]{ieeeconf}

\IEEEoverridecommandlockouts                              % This command is only needed if 
\usepackage{xcolor}
\usepackage{graphics} % for pdf, bitmapped graphics files
\usepackage{epsfig} % for postscript graphics files
\usepackage{mathptmx} % assumes new font selection scheme installed
\usepackage{amsmath} % assumes amsmath package installed
\usepackage{amssymb}  % assumes amsmath package installed
\usepackage{textgreek}
\usepackage{hyperref}
\usepackage{stfloats}
\usepackage{flushend}
\usepackage{subfigure}

\usepackage[belowskip=-10pt,aboveskip=12pt]{caption}

\title{\LARGE \bf Design, Integration and Sea Trials of 3D Printed Unmanned Aerial Vehicle and Unmanned Surface Vehicle for Cooperative Missions}

\author{Hanlin Niu, Ze Ji, Pietro Liguori, Hujun Yin, and Joaquin Carrasco% <-this % stops a space
\thanks{*This work was partially supported by EPSRC project No.EP/S03286X/1. We would like to thank the group students of the Mechatronics module for their contributions to the project at Cardiff University. \textit{(Corresponding author: Hanlin Niu)}}
\thanks{H. Niu, H. Yin and J. Carrasco are with the Department of Electrical \& Electronic Engineering, The University of Manchester, Manchester, UK.
        {\{\tt\small hanlin.niu@manchester.ac.uk}\}}%
\thanks{Z. Ji and P. Liguori are with the School of Engineering, Cardiff University, Cardiff, UK.}%%

}

\begin{document}

\maketitle
\thispagestyle{empty}
\pagestyle{empty}

%%%%%%%%%%%%%%%%%%%%%%%%%%%%%%%%%%%%%%%%%%%%%%%%%%%%%%%%%%%%%%%%%%%%%%%%%%%%%%%%
\begin{abstract}

In recent years, Unmanned Surface Vehicles (USV) have been extensively deployed for maritime applications. However, USV has a limited detection range with sensor installed at the same elevation with the targets. In this research, we propose a cooperative Unmanned Aerial Vehicle - Unmanned Surface Vehicle (UAV-USV) platform to improve the detection range of USV. A floatable and waterproof UAV is designed and 3D printed, which allows it to land on the sea. A catamaran USV and landing platform are also developed. To land UAV on the USV precisely in various lighting conditions, IR beacon detector and IR beacon are implemented on the UAV and USV, respectively. Finally, a two-phase UAV precise landing method, USV control algorithm and USV path following algorithm are proposed and tested.
%, when transferred from the simulation environment to the real world.
\end{abstract}

\section{Introduction}

Over the past decade, advances in computing devices and navigation sensors accelerated the development and the range of missions USVs can undertake. USV are extensively utilised for applications such as seabed mapping, marine animals tracking, and maintenance of offshore facilities such as wind farms or oil pipelines. Compared to manned vehicles, a USV could be deployed for dangerous and boring tasks continuously in shallow and restricted waters \cite{ma2018multi}. As a USV could carry heavy payload, it can install multiple energy resources to achieve long endurance performance \cite{niu2018energy}. However, USV can only perceive dynamic sea condition in a short range. By contrast, UAV is capable of detecting large-scale sea condition at a high altitude and has better maneuverability than USV. However, UAV has limited flight endurance due to its limited battery capacity. By developing a system that can allow for a UAV to autonomously land on the USV, the flight range of UAVs will be considerably increased. UAV could also detect the USV surrounding environment at a high altitude \cite{xiao2017uav}, which improves the detection range of USV. Moreover, UAV can be used as a communication relay between USV and ground station, which enlarges the communication range of USV. This research proposes an affordable system that allows the UAV and USV work collaboratively and autonomously for making uses of the strengths and diminishing the weaknesses of both UAVs and USVs.

\begin{figure} [!t]
  \begin{center}
    \includegraphics[width=3.4in]{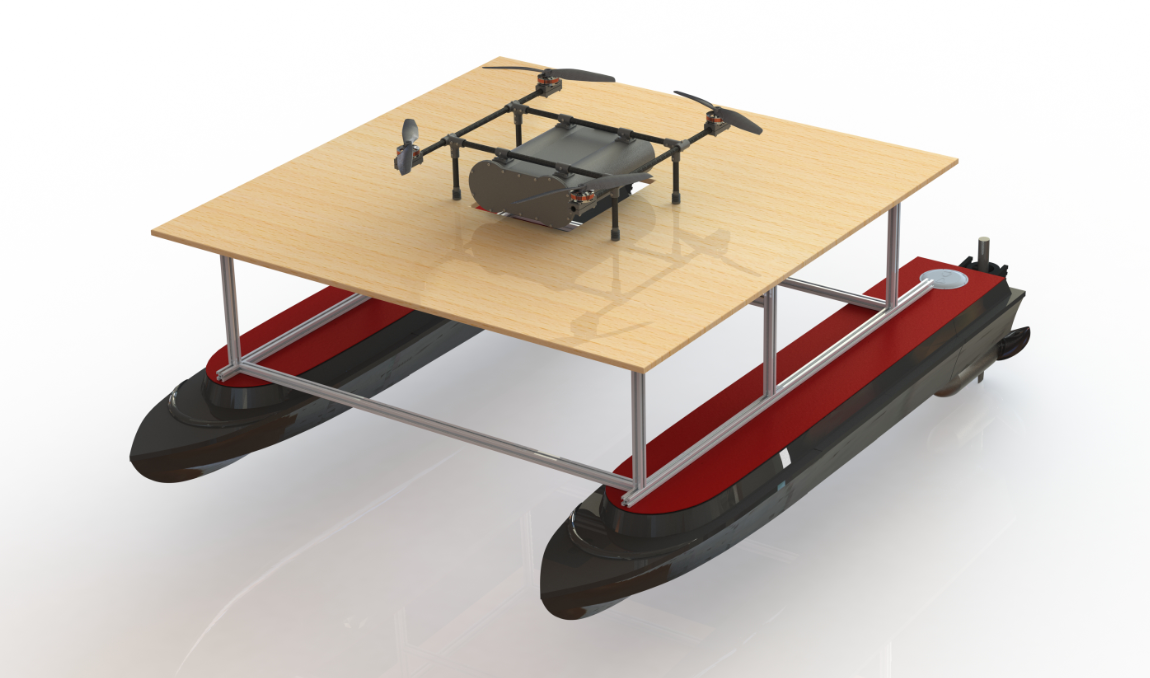}
    \caption{3D printed UAV and USV.}
    \label{fig1}
  \end{center}
\end{figure}

\section{UAV and USV Design}

\subsection{UAV Design}
\begin{figure}[h]
  \begin{center}
    \includegraphics[width=3.25in,height=1.6in]{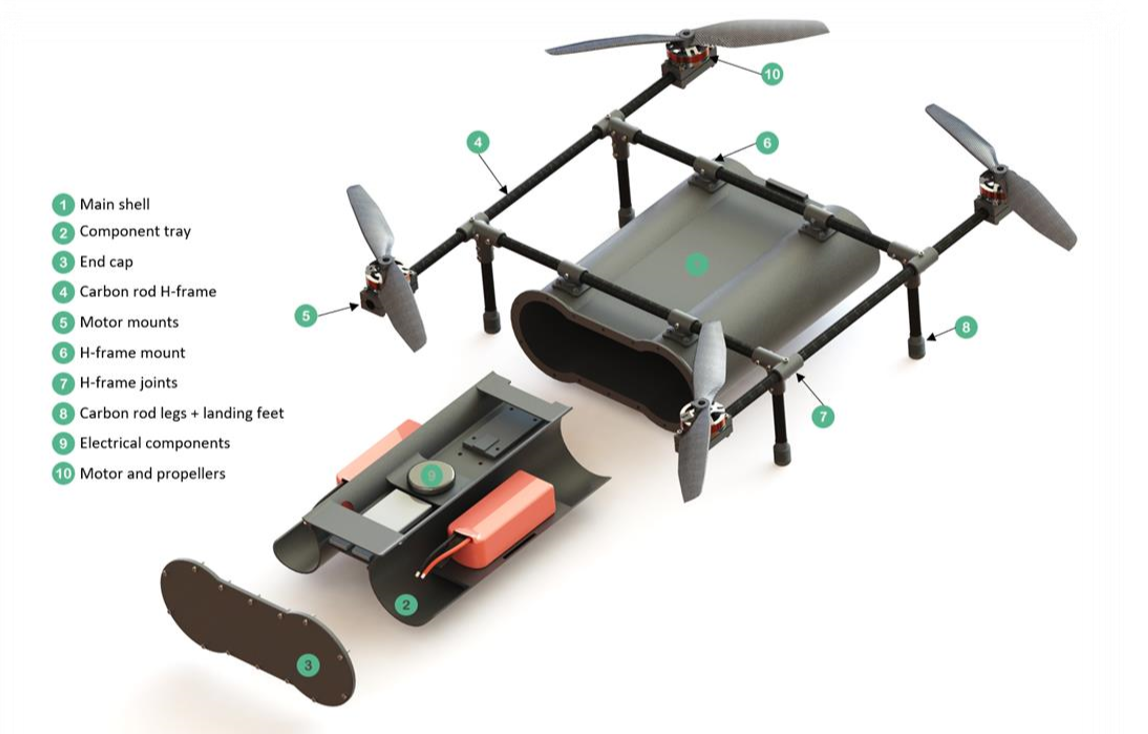}
    \caption{CAD design of waterproof UAV.}
    \label{fig21}
  \end{center}
\end{figure}

To protect the electronic components from sea water, we designed a waterproof UAV, which was 3D printed using Selective Laser Sintering (SLS) printer and PA2200 printing material, as shown in Fig.~\ref{fig21}. The volume of the shell is 0.004006 cubic meters and its buoyancy can support a mass of $4.11kg$, which enables the UAV to float on water. To enable the UAV to localize the USV position precisely under various lighting conditions, we used an IR beacon detector on the bottom of UAV. The IR beacon detector was designed to filter most light noises and only detect the light within certain frequency spectrum emitted by the IR beacon, which was installed on the USV landing platform. The GPS, autopilot, battery and IR beacon detector were installed on the component tray, which is inside of the shell, as shown in Fig.~\ref{fig21}.

\subsection{Design of the USV and its Landing Platform}

\begin{figure}[hbt!]
\centering
\subfigure[]{\label{fig24}
\includegraphics[width=0.23\textwidth,height=0.16\textwidth]{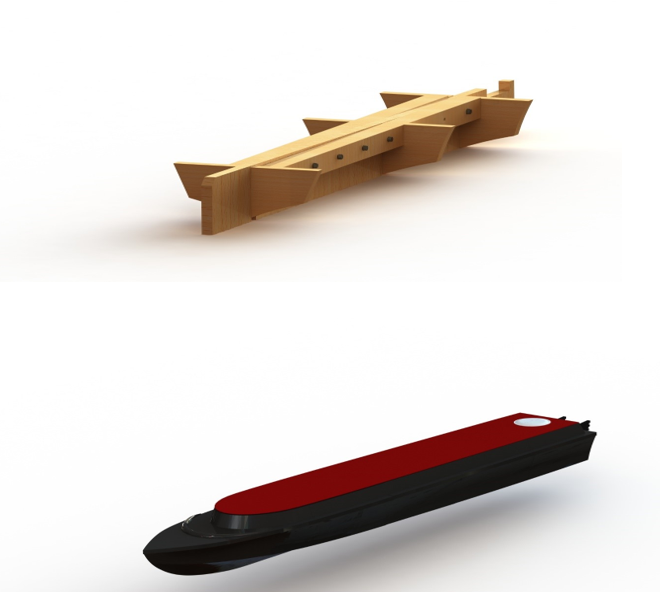}}
\subfigure[]{\label{fig25}
\includegraphics[width=0.23\textwidth,height=0.16\textwidth]{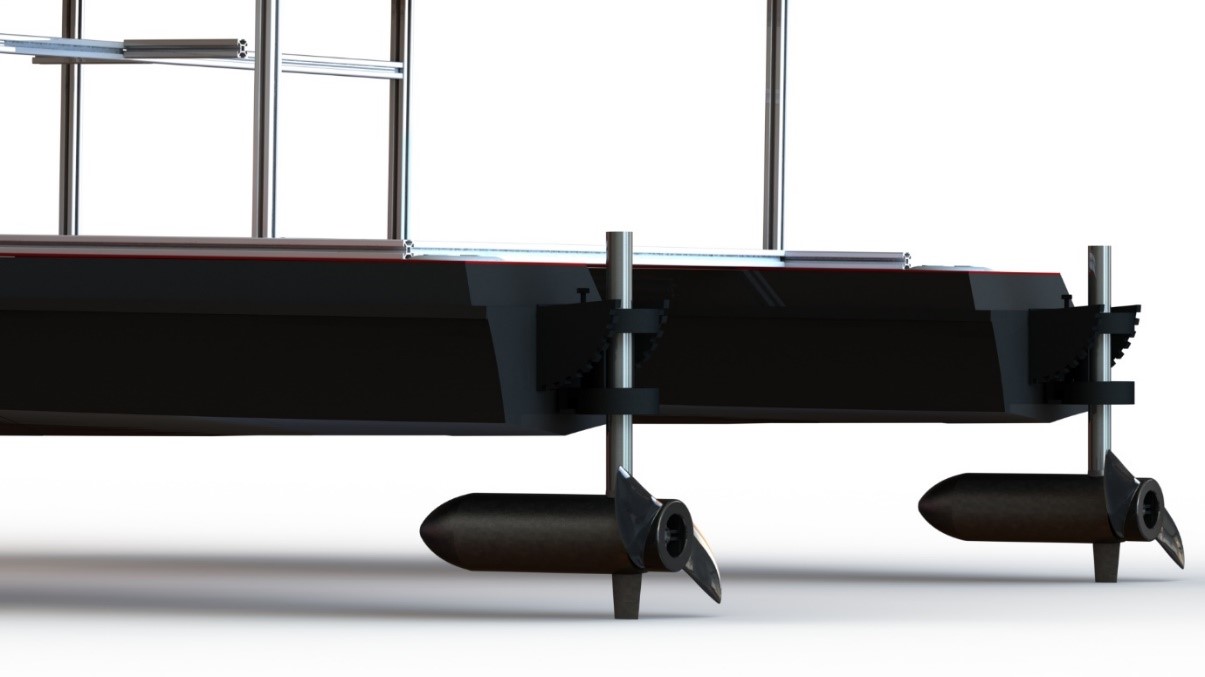}}
\subfigure[]{\label{fig26}
\includegraphics[width=0.23\textwidth,height=0.16\textwidth]{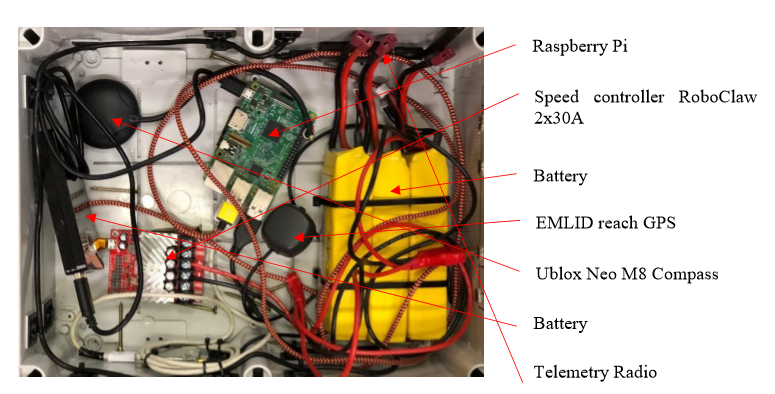}}
\subfigure[]{\label{fig27}
\includegraphics[width=0.23\textwidth,height=0.16\textwidth]{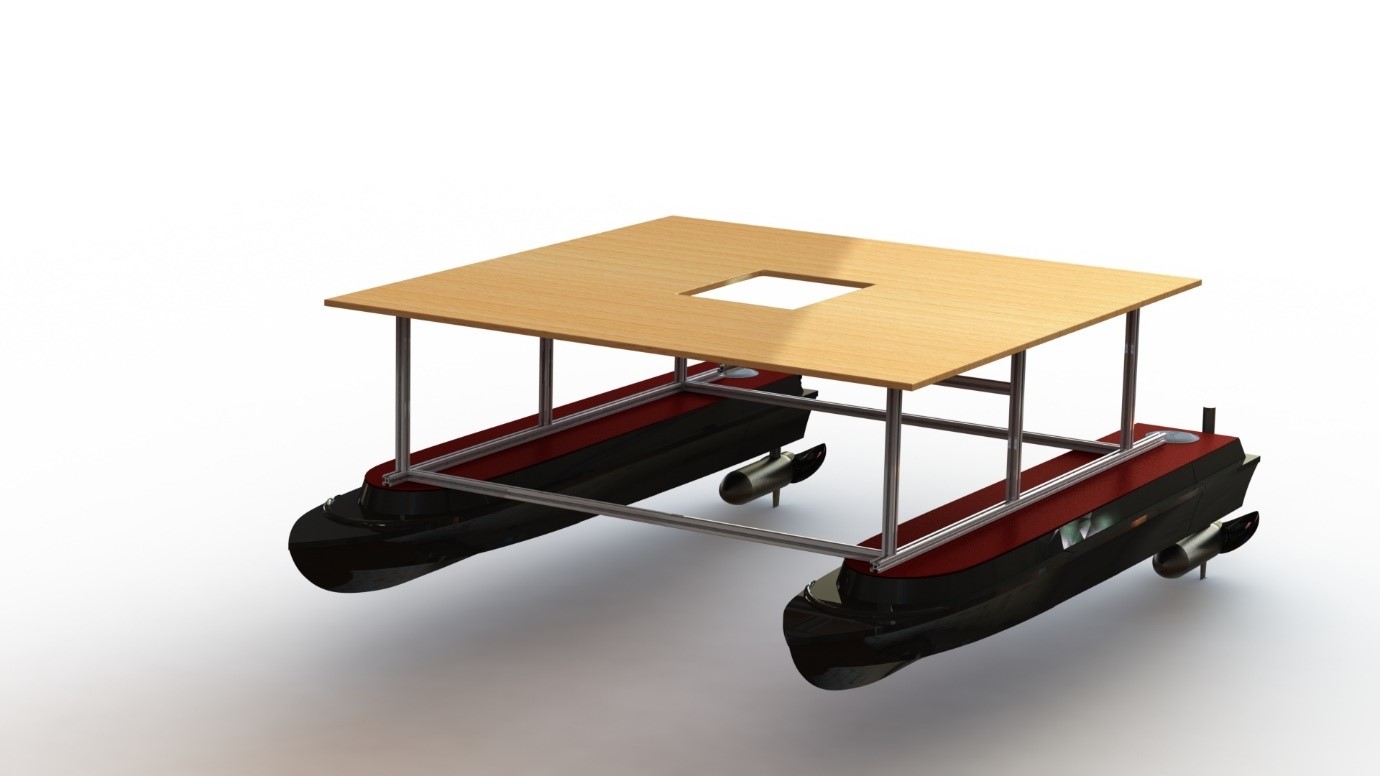}}
\caption{Catamaran USV: (a) Wooden keel and 3D printed hull (b) Propulsion system of USV (c) USV control box (d) Catamaran USV and landing platform}
\end{figure}

To improve the stability of the landing platform, a catamaran USV was designed. The CAD design of hull and wooden keel are shown in Fig.~\ref{fig24}. It is calculated to carry a payload of up to $32.5kg$ per hull when fully submerged. Polyethylene Terephthalate Glycol (PETG) filament was used to print hulls since it can produce strong parts with slight flexibility and has a relatively low shrinkage ratio compared to other filament types such as Acrylonitrile Butadiene Styrene (ABS) and Polylactic Acid (PLA). The final hull dimensions was $1.655m \times 0.301m \times 0.155m$ and the total mass was 8.1kg per hull. MotorGuide R3-30 trolling motors were selected and fitted via modified mounts that allowed the motors position adjustable, as shown in Fig.~\ref{fig25}. The maximum thrust of motors can be up to $133.8N$. The weight of each hull with fully assembled motors and deck was $17.4kg$.
All control units of the USV, including raspberry pi, speed controller RoboClaw 2x30A, Battery, Telemetry Radio, Ublox Neo M8 compass and Emlid reach GPS, were installed in a waterproof box, as shown in Fig.~\ref{fig26}, which was located in the centre of the landing platform with aluminium frame supports, as shown in Fig.~\ref{fig27}.

\section{Experiments and Discussion}
\subsection{Two-Phase Precise Landing Test}
\begin{figure}[t]
\centering
\subfigure[]{\label{fig41}
\includegraphics[width=3.0in]{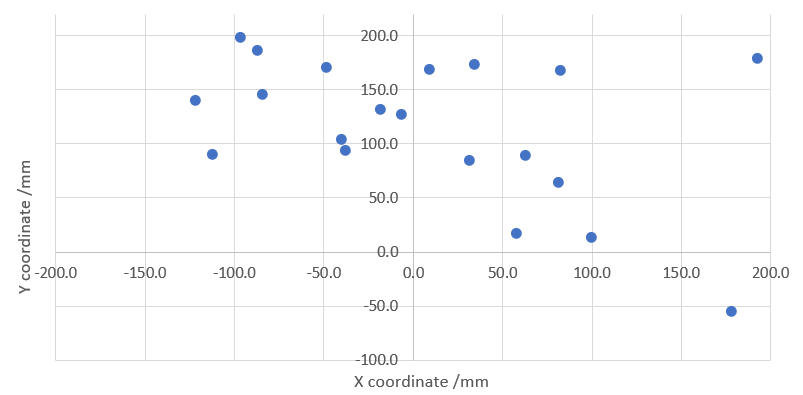}}
\subfigure[]{\label{fig42}
\includegraphics[width=2.8in]{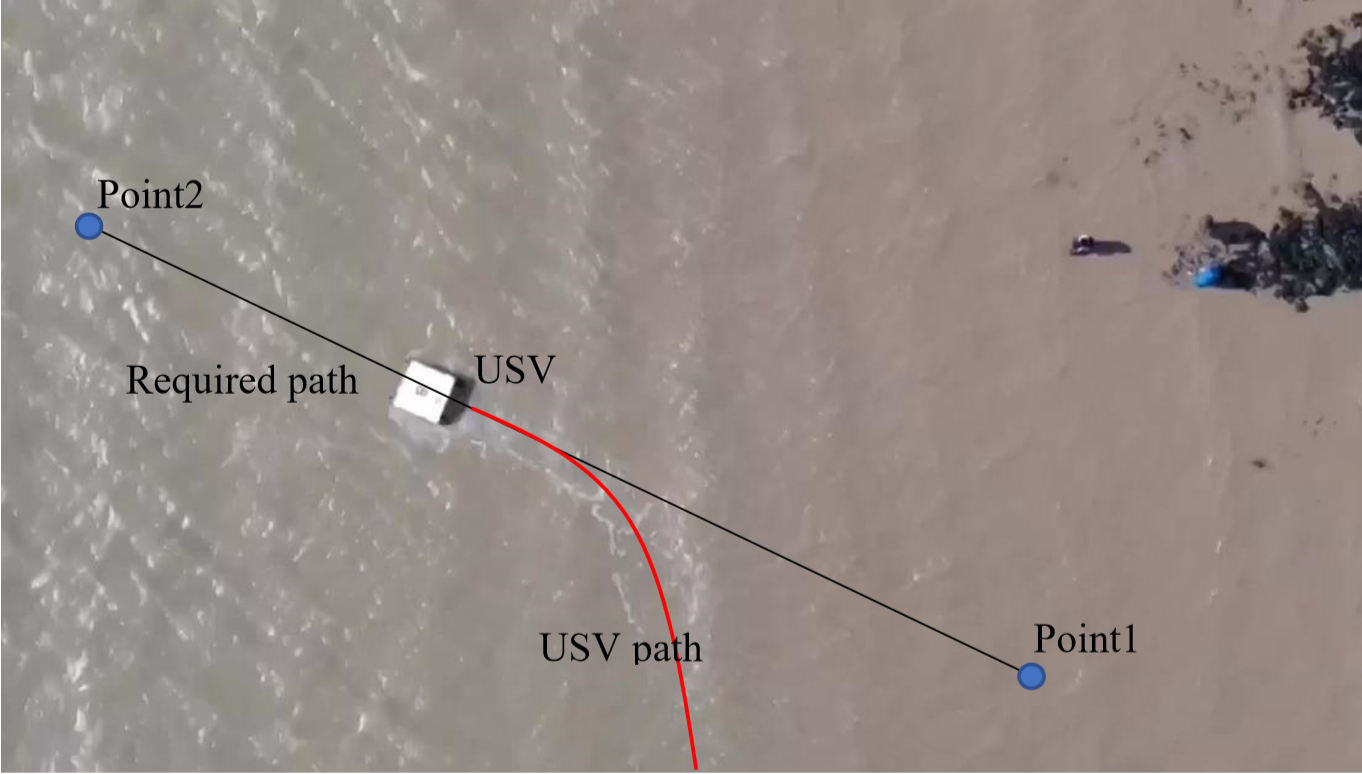}}

\caption{UAV and USV experiments: (a) Accuracy results of twenty landing tests (b) Aerial view of USV control algorithm and path following algorithm test}
\end{figure}

A two-phase landing method was proposed to land the UAV on the USV precisely using standard GPS and IR beacon. At first, the UAV navigates to the USV location using USV GPS data. Standard GPS data are not accurate enough to allow accurate landing on the platform due to the limited dimension. When the UAV is above the USV and localizes the IR beacon, precise landing using IR beacon will be activated. The accuracy of precise landing is evaluated using twenty tests. As shown in Fig.~\ref{fig41}, the relative position between the IR beacon detector on the UAV and the IR beacon in the centre of the USV landing platform satisfies: $\Delta X < 20cm$, $\Delta Y < 20cm$. Note that the offset along y axis was caused by the wind during the tests. As the landing platform dimension is $1.2m \times 1m$ that is larger than the landing derivation, the UAV could land on the USV safely. 

% \begin{figure}
%   \begin{center}
%     \includegraphics[width=3.4in]{Landingprecision.PNG}
%     \caption{Accuracy results of twenty landing tests}
%     \label{fig4}
%   \end{center}
% \end{figure}

\subsection{Control and Path Following Algorithm of USV}
PI controller and PD controller were implemented for speed control and heading control \cite{niu2016efficient}, respectively. To navigate the USV to follow the pre-defined path precisely, carrot chasing path following algorithm \cite{niu2016efficient} was implemented on USV for generating required heading and speed command. As shown in Fig.~\ref{fig42}, two waypoints were pre-defined by ground station. The USV was able to follow the pre-defined path, represented by the line between point1 and point2.

% \begin{figure}
%   \begin{center}
%     \includegraphics[width=3.2in]{carrotchasing.PNG}
%     \caption{Aerial view of USV control algorithm and path following algorithm test}
%     \label{fig5}
%   \end{center}
% \end{figure}

%\section*{Acknowledgment}
%This work was supported by the EPSRC project "Digital twin-based Bilateral Teleautonomous System for Nuclear Remote Operation"(EP/S03286X/1).

\bibliographystyle{IEEEtran}
\bibliography{IEEEexample}

\end{document}